\documentclass[letterpaper, 10 pt, conference]{ieeeconf}  %

\IEEEoverridecommandlockouts                              %

\usepackage{graphicx}
\usepackage{epsfig} %
\usepackage{url}
\usepackage[acronym]{glossaries}
\usepackage[toc=false,abbreviations,symbols,nogroupskip,nonumberlist]{glossaries-extra}
\usepackage{booktabs} %
\usepackage{amsfonts} %
\usepackage{amsmath}
\usepackage{amssymb}
\usepackage{nicefrac} %
\usepackage{microtype} %
\usepackage{doi}
\usepackage[caption=false, font=footnotesize]{subfig}
\usepackage[dvipsnames, table]{xcolor} %
\usepackage{multirow}
\usepackage{floatrow}
\usepackage{blindtext}
\usepackage{bm} %
\usepackage{upgreek} %
\usepackage{listings} %
\usepackage{tikz}
\usepackage{balance} %
\usepackage{soul}
\usetikzlibrary{patterns} %
\usepackage{dblfloatfix}
\usepackage{cite} 
\usepackage{makecell}
\usepackage[symbol*]{footmisc}
\usepackage{wrapfig}
\usepackage{mwe} %
\usepackage{hyperref}
\usepackage{cuted}
\usepackage{capt-of}

\DeclareMathOperator*{\avg}{avg}
\hypersetup{
    colorlinks,
    linkcolor={black},
    citecolor={black},
    urlcolor={blue}
}

\graphicspath{{images/}}

\makeatletter
\def\input@path{{chapters/}}
\makeatother

\title{\LARGE \bf
CrazyMARL: Decentralized Direct Motor Control Policies for Cooperative Aerial Transport of Cable-Suspended Payloads
}

\author{%
Viktor~Lorentz\textsuperscript{1}, Khaled~Wahba\textsuperscript{1}, Sayantan~Auddy\textsuperscript{1}, Marc~Toussaint\textsuperscript{1,2}, and Wolfgang~Hönig\textsuperscript{1,2}%
\thanks{Corresponding author: \texttt{viktor.lorentz@hhi.fraunhofer.de}}
\thanks{\textsuperscript{1}Technische Universität Berlin, Berlin, Germany.}%
\thanks{\textsuperscript{2}Robotics Institute Germany (RIG), Germany.}%
\thanks{This work was supported by the German Federal Ministry of Research, Technology and Space (BMFTR) under the Robotics Institute Germany (RIG), and the Deutsche Forschungsgemeinschaft (DFG) - 448549715.}
}

\makeglossaries[abbreviations]
\newacronym{api}{API}{Application Programming Interface}
\newabbreviation{ctbr}{CTBR}{Collective Thrust and Body Rates}
\newabbreviation{ctde}{CTDE}{Centralized Training with Decentralized Execution}
\newabbreviation{dec-pomdp}{Dec-POMDP}{Decentralized Partially Observable Markov Decision Process}
\newabbreviation{ekf}{EKF}{Extended Kalman Filter}
\newabbreviation{gae}{GAE}{Generalized Advantage Estimation}
\newabbreviation{imu}{IMU}{Inertial Measurement Unit}
\newabbreviation{ippo}{IPPO}{Independent Proximal Policy Optimization}
\newabbreviation{lv}{LV}{Linear Velocity and Yaw Rate}
\newabbreviation{mappo}{MAPPO}{Multi-Agent Proximal Policy Optimization}
\newabbreviation{marl}{MARL}{Multi-Agent Reinforcement Learning}
\newabbreviation{mdp}{MDP}{Markov Decision Process}
\newabbreviation{mlp}{MLP}{Multilayer Perceptron}
\newabbreviation{mocap}{MoCap}{Motion Capture}
\newabbreviation{mpc}{MPC}{Model Predictive Control}
\newabbreviation{ppo}{PPO}{Proximal Policy Optimization}
\newabbreviation{pwm}{PWM}{Pulse Width Modulation}
\newabbreviation{rl}{RL}{Reinforcement Learning}
\newabbreviation{sgd}{SGD}{Stochastic Gradient Descent}
\newabbreviation{srt}{SRT}{Single-Rotor Thrust}
\newabbreviation{uav}{UAV}{Unmanned Aerial Vehicle}
\newabbreviation{dr}{DR}{Domain Randomization}

\begin{document}

\maketitle
\begin{strip}
\vspace{-5.5\baselineskip}
  \centering
  \includegraphics[width=\textwidth]{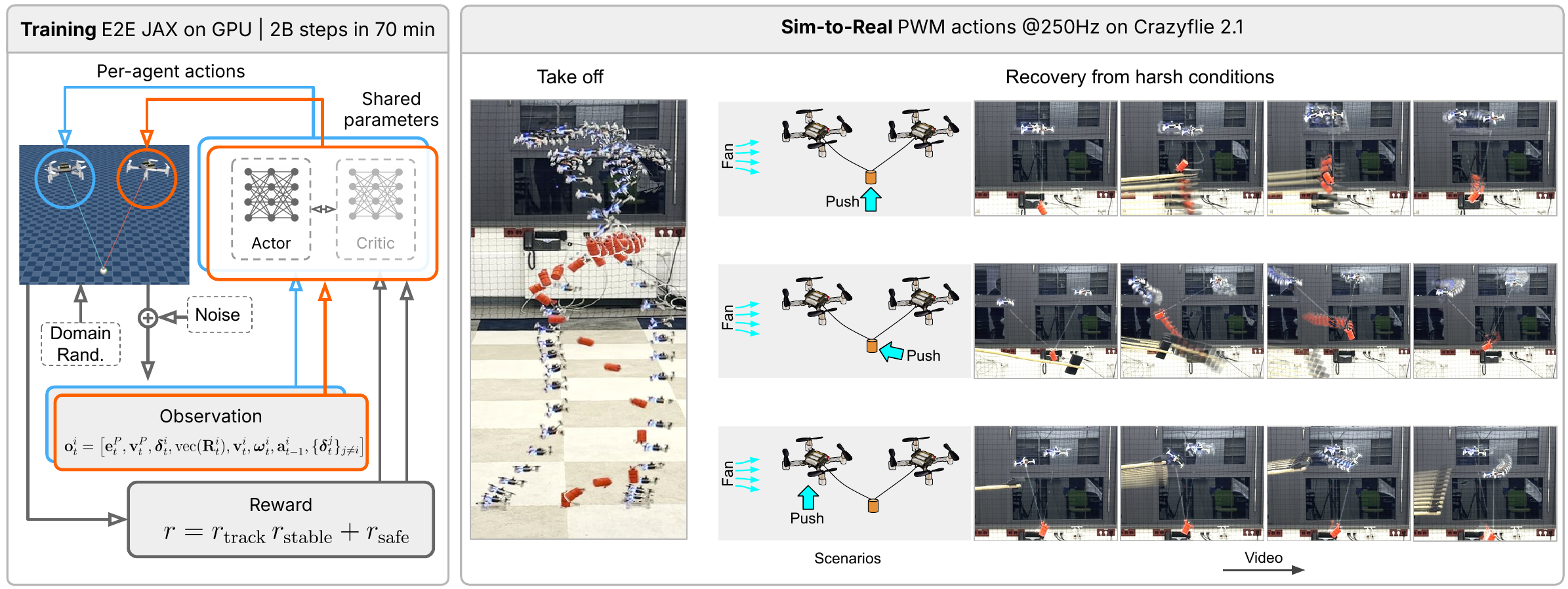}
  \captionof{figure}{Overview of our approach. (Left panel) Training in Mujoco with end-to-end JAX and domain randomization. \gls{ippo} with shared parameters maps each per agent observation $o_t^i=[\,e_t^P, v_t^P, \delta_t^i, \mathrm{vec}(R_t^i), v_t^i, \omega_t^i, a_{t-1}^i, \{\delta_t^j\}_{j\neq i}\,]$ to motor commands under a reward that combines tracking, stability, and safety. The local observation packs payload error and velocity, the quadrotors' relative position, orientation, and velocities, its last action, and for coordination, the relative positions of teammates. 
  (Right panel) Sim-to-Real on the Crazyflie~2.1 robot, where the same decentralized policy runs on two quadrotors at $250\,\mathrm{Hz}$ and outputs direct \gls{pwm}. The sequences show autonomous takeoff and recovery scenarios of a cable-suspended payload under strong pushes and wind (shown with cyan-colored arrows) with an average speed of 3.5~m/s. Video frames progress from left to right.}
  \label{fig:overview}    
\end{strip}

\begin{abstract}
Collaborative transportation of cable-suspended payloads by teams of \glspl{uav} has the potential to enhance payload capacity, adapt to different payload shapes, and provide built-in compliance, making it attractive for applications ranging from disaster relief to precision logistics. However, multi-\gls{uav} coordination under disturbances, nonlinear payload dynamics, and slack–taut cable modes remains a challenging control problem. To our knowledge, no prior work has addressed these cable mode transitions in the multi-\gls{uav} context, instead relying on simplifying rigid-link assumptions. We propose \textit{CrazyMARL},
a decentralized \gls{rl} framework for multi-\gls{uav} cable-suspended payload transport. Simulation results demonstrate that the learned policies can outperform classical decentralized controllers in terms of disturbance rejection and tracking precision, achieving an 80\% recovery rate from harsh conditions compared to 44\% for the baseline method. 
We also achieve successful zero-shot sim-to-real transfer and demonstrate that our policies are highly robust under harsh conditions, including wind, random external disturbances, and transitions between slack and taut cable dynamics.
This work paves the way for autonomous, resilient \gls{uav} teams capable of executing complex payload missions in unstructured environments. 
Code and videos can be found on the website: \texttt{
https://imrclab.github.io/CrazyMARL}.

\end{abstract}

\section{Introduction}
\glsresetall
\glspl{uav} have moved from research prototypes to widely deployed tools in civil and industrial settings, including disaster response, agriculture, logistics, and inspection \cite{Idrissi2022AROA,Lyu2023UnmannedAVA}. Among emerging applications, collaborative cable-suspended payload transport is compelling: cables are lightweight, adaptable to irregular loads, and introduce compliance that attenuates vibration. However, multi-\gls{uav} transport of a shared suspended payload is challenging due to pendulum dynamics, tension coupling, contact events, and disturbances. 

There are two well-established paradigms to tackle control in robotics: model-based and learning-based methods. Classical model-based controllers offer stability guarantees and well-founded design principles, but struggle with modeling errors and scalability. Centralized coordination can provide optimal solutions but incurs bottlenecks and single points of failure, while decentralized schemes can tackle scalability but may lack performance guarantees~\cite{batra_decentralized_2022,estevez_review_2024}. \gls{rl}, in contrast, has emerged as a complementary paradigm that can provide adaptivity in scenarios with complex dynamics or incomplete models.
It can directly learn control from interaction and improve robustness to unmodeled effects and disturbances. 
Recent results demonstrate strong performance for single and multi \gls{uav} tasks such as agile flights under challenging conditions \cite{kaufmann_champion-level_2023,Eschmann2024}. 

Building on this progress, we leverage \gls{rl} to address the decentralized control problem of multi-\gls{uav} payload transport with hybrid cable dynamics that capture taut–slack transitions. 
To the best of our knowledge, this is the first use of \gls{rl} for controlling multiple robots with constrained onboard microcontrollers operating near their actuation limits. 
Our objectives are to stabilize the suspended payload under external disturbances, manage cable mode switching, and safely distribute forces among the robots without collisions.
Fig.~\ref{fig:overview} illustrates the training pipeline, deployed system, and shows hardware demonstrations with two robots transporting a payload, including robust recovery from harsh conditions and strong disturbance rejection.

We implement our method on the Crazyflie 2.1 research platform, which is widely used for cooperative transport and \gls{uav} control studies \cite{wahba_kinodynamic_2024,huang_collision_2024}.
The target platform has a low thrust-to-weight ratio of 1.4, which makes operation close to the motor limits much more likely than on other platforms such as high-end racing multirotors.
Our method allows each robot to execute its own policy onboard at 250~Hz and relies on local state estimation and relative pose information. 
The policy output maps directly to motor \gls{pwm} at high frequency, without relying on cascaded low-level controllers. This enables us to operate close to the actuation limits which is particularly important for agile flights and disturbance rejection. 

On the learning side, we leverage high throughput training with large scale parallelized simulation and extensive domain randomization to improve robustness and shorten iteration time. 
The resulting controller remains fully decentralized with a small computational footprint and no need for inter-robot communication, which simplifies deployment and improves real world performance.

In summary, our main contributions are:
\begin{itemize}
  \item \textbf{End-to-end decentralized \gls{marl} for multiple quadrotors carrying a cable-suspended payload.} We train a fully decentralized policy with direct motor PWM commands (no low-level cascades) for cable-suspended payload transport.
  \item \textbf{Empirical evaluation in simulation and hardware.} We validate on Crazyflie 2.1 platform. In wind trials with a mean wind speed of 3.5~m/s the policy maintains stable formations and recovers from large external disturbances.
  
  \item \textbf{High-throughput simulation.} A GPU-parallelized JAX/MJX pipeline captures cable–payload–robot interactions in contact-rich scenarios, providing a foundation for future aerial manipulation and related tasks.
\end{itemize}

\section{Related Work}

We review control strategies for multirotor \glspl{uav} transporting cable-suspended payloads, covering classical model-based methods as well as learning-based approaches for both single- and multi-agent coordination. 
For a comprehensive survey of aerial cable transport, see~\cite{estevez_review_2024}.

\subsection{Traditional Model-based Approaches}

Model-based control approaches for aerial payload transport include centralized cascaded geometric controllers that provide stability guarantees for payload and manipulation control~\cite{sreenath_dynamics_2013}, as well as decentralized controllers that exploit internal cable tension for quasi-static attitude stabilization~\cite{tognon_aerial_2018}. 
However, these methods rely on noisy payload acceleration as a feedback signal and model cables as rigid rods, which limit the applicability in agile scenarios. 

Centralized and decentralized nonlinear model predictive control (NMPC) have advanced multi-\gls{uav} payload manipulation without acceleration feedback and can incorporate high-level objectives (e.g., obstacle collision avoidance)~\cite{ sun_nonlinear_2023,de2025distributed}. 
However, centralized NMPC is computationally expensive and suffers when scaled to large teams, while decentralized NMPC depends on iterative inter-robot communication and can suffer from deadlocks. Both approaches still adopt the rigid-rod cable assumption inherited from reactive controllers.  

The controllers devised by the previous methods adopt simplified models which restrict control performance to quasi-static regimes. Thus, other works have employed the full system dynamics in offline motion planning through optimization-based planners to account for more accurate models and enable agile maneuver planning~\cite{wahba2025pc, wahba_kinodynamic_2024,Wang2025SafeAA,sun2025agile}. Nevertheless, despite considering the full dynamics, these approaches still rely on the rigid-rod cable assumption.

To overcome the limitations of the rigid-rod assumption, more accurate cable models have been incorporated by switching the dynamics that explicitly capture the taut–slack transitions~\cite{wang2024impact, recalde2025hpc}. NMPC formulations with such hybrid models enable motions unattainable under rigid-rod assumptions but have so far been limited to single-\gls{uav} systems. Extending them to cooperative multi-robot manipulation remains challenging due to the added complexity of modeling, optimization, and coordination.

In contrast, our work explores reinforcement learning (RL) as a promising direction for achieving agile maneuvers and high robustness in multi-\gls{uav} payload transport while accounting for realistic cable dynamics.
\subsection{Reinforcement Learning-based Approaches}

Early application of \gls{rl} to \gls{uav} control focused on single-agent scenarios, and showed that RL agents trained with model-free \gls{rl} algorithms perform on par with or better than classical controllers~\cite{Koch2018ReinforcementLF}. 
Later works have demonstrated the effectiveness of \gls{rl}-trained \glspl{uav} in handling harsh initial conditions, executing aggressive maneuvers, and operating near the limits of their dynamic capabilities~\cite{Song2023ReachingTL, xing_multi-task_2024}. 
Other approaches have extended single-\gls{uav} control to payload transport and aerial manipulation, where RL-based controllers have also proven effective in adapting to unknown payload dynamics, maintaining stability and rejecting payload disturbances~\cite{hua_new_2022}.
Notably, \cite{cao2025flare} is capable of mode-switching and handling flexible cables for single \glspl{uav}.
However, compared to the single-\gls{uav}-payload-transportation problem, our current work on multi-\gls{uav} collaborative payload transport is considerably more challenging due to the coupling between vehicles, the need for precise coordination to regulate cable tensions, and the heightened risk of collisions.

\glsreset{marl}\gls{marl} has emerged as a powerful framework for cooperative tasks.
Some \Gls{marl} approaches employ \gls{ctde} by using a shared critic, such as \cite{Lowe2017MultiAgentAF, yu_surprising_2022}. Others, such as \gls{ippo}~\cite{witt_is_2020}, follow a completely decentralized regime. A decentralized scheme such as \gls{ippo} avoids scalability and communication overheads and is thus utilized in our work.
Many works have successfully used \gls{marl}, particularly methods such as \gls{mappo} and \gls{ippo}, in real-world, multi-robot collaborative tasks~\cite{Pandit2024LearningDM,  Chen2025DecentralizedNO}.
\gls{marl} strategies have also achieved success in swarm coordination, collaborative pursuit and evasion, and obstacle avoidance~\cite{huang_collision_2024, zhao_deep_2024}.

In the area of payload transportation, the adaptability and disturbance rejection capabilities of single-\gls{uav} methods have naturally led to multi-\gls{uav} extensions such as \cite{Lin2024PayloadTW, Estevez2024Reinforcement, xu_omnidrones_2024}. However, unlike the centralized approach presented in \cite{Lin2024PayloadTW}, we adopt a decentralized approach, thereby avoiding communication and scalability overheads.
While \cite{xu_omnidrones_2024} considers only rigid-link payloads and \cite{Estevez2024Reinforcement} assumes the payload cables are always taut, we make no such assumptions and model payload links with realistic flexible cables. Neither of these works reports transfer to the real world, whereas we achieve successful zero-shot sim2real transfer and demonstrate the robustness and agility of our robots under harsh real-world conditions. 
The closest to our current work is the approach from~\cite{zeng2025decentralized}, which also uses a decentralized training scheme and showcases robustness under harsh real-world settings. 
However, unlike our fully decentralized \gls{ippo}-based approach, this work uses the \gls{ctde}-based \gls{mappo} algorithm. It relies on a low-level controller, whereas we do not, and thanks to our highly parallelized training setup, our training speed is about ten times faster. In addition, \cite{zeng2025decentralized} still retains the rigid-rod assumption, where our work focuses on exploiting the hybrid cable model to achieve agile maneuvers and recovery from harsh configurations.

\section{Method: MARL for Cooperative Payload Transport}

We train decentralized \gls{marl} policies that enable $Q$ Crazyflie quadrotors to transport a cable suspended payload. The task is modeled as a \gls{dec-pomdp}
$
(\mathcal{Q},\mathcal{S},\{\mathcal{A}^i\},P,r,\{\Omega^i\},O,\rho_0,\gamma),
$
where $\mathcal{Q}=\{1,\dots,Q\}$ is the agent index set, $\mathcal{S}$ is the state space, $\mathcal{A}^i=[-1,1]^4$ is the action space of agent $i$, $P$ is the transition kernel, $r$ is the shared reward, $\Omega^i$ is the local observation space of agent $i$, $O$ is the observation function, $\rho_0$ is the reset distribution, and $\gamma\in[0,1)$ is the discount.

The global state at time $t$ is
\begin{equation}
\mathbf{s}_t=\Bigl(\{\mathbf{p}^i_t,\mathbf{v}^i_t,\mathbf{R}^i_t,\boldsymbol{\omega}^i_t\}_{i=1}^Q,\,\mathbf{p}^P_t,\mathbf{v}^P_t\Bigr),
\end{equation}
with $\mathbf{p}^i_t\in\mathbb{R}^3$ and $\mathbf{v}^i_t\in\mathbb{R}^3$ the position and linear velocity of agent $i$, $\mathbf{R}^i_t\in\mathrm{SO}(3)$ its attitude, $\boldsymbol{\omega}^i_t\in\mathbb{R}^3$ its body rates, and $\mathbf{p}^P_t,\mathbf{v}^P_t\in\mathbb{R}^3$ are the payload position and velocity.

Each agent produces a normalized motor command $\mathbf{a}^i_t\in[-1,1]^4$. We write $\mathbf{a}_t$ for the joint action, the concatenation of $\{\mathbf{a}^i_t\}_{i=1}^Q$. The team receives a shared reward $r(\mathbf{s}_t,\mathbf{a}_t)$. The observation spaces are $\{\Omega^i\}$ and the observation function $O$ yields local observations $\mathbf{o}^i_t$ from $\mathbf{s}_t$ as defined below. The initial state is sampled from $\rho_0$ through the randomized reset distribution. The objective is

\begin{equation}
\max_{\theta}\;J(\theta)=\mathbb{E}\!\left[\sum_{t=0}^{\infty}\gamma^t\,r(\mathbf{s}_t,\mathbf{a}_t)\right],
\quad
\mathbf{a}^i_t\sim\pi_{\theta}(\cdot\mid \mathbf{o}^i_t,\mathbf{a}^i_{t-1}).
\end{equation}
We use \gls{ippo} with parameter sharing. A single actor and critic are trained on data from all agents and the critic conditions only on $\mathbf{o}^i_t$ and $\mathbf{a}^i_{t-1}$. Training is centralized by shared parameters and execution is decentralized, following a \gls{ctde} paradigm without privileged information.
To mitigate partial observability, improve stability, and ease sim-to-real transfer, we augment each local observation with the previous action and encourage smooth commands during training.

\subsection{CrazyMARL Framework}
We propose CrazyMARL, an end-to-end JAX-based pipeline that couples MuJoCo MJX with JaxMARL algorithms \cite{todorov2012mujoco,flair2023jaxmarl}. It is designed for training coordinated behaviors with multiple quadrotors. We specialize it for use with the Crazyflie research platform carrying a cable-suspended payload, but it can be easily reparameterized for other multirotors and scenarios. MJX allows us to run thousands of environments in parallel on GPU. Our tasks cover single-robot hover/tracking and multi-robot cable-suspended transport with configurable payloads and cables (MuJoCo tendons). We focus on disturbance rejection and hovering under harsh conditions.

\subsection{Observations}
We form a global observation
\begin{equation}
\mathbf{o}_t=\bigl[\mathbf{e}^P_t,\mathbf{v}^P_t,\{\boldsymbol{\delta}^i_t,\mathrm{vec}(\mathbf{R}^i_t),\mathbf{v}^i_t,\boldsymbol{\omega}^i_t,\mathbf{a}^i_{t-1}\}_{i=1}^Q\bigr],
\label{eq:obs_global}
\end{equation}
where $\mathbf{e}^P_t=\mathbf{p}^P_{\mathrm{des},t}-\mathbf{p}^P_t$, $\boldsymbol{\delta}^i_t=\mathbf{p}^i_t-\mathbf{p}^P_t$, and
$\mathrm{vec}(\mathbf{R}^i_t)$ denotes the column vector obtained by stacking the columns of $\mathbf{R}^i_t$.
For decentralized execution, agent $i$ observes its local state and the agents' positions relative to the payload.
\begin{equation}
\mathbf{o}^i_t=\bigl[\mathbf{e}^P_t,\mathbf{v}^P_t,\boldsymbol{\delta}^i_t,\mathrm{vec}(\mathbf{R}^i_t),\mathbf{v}^i_t,\boldsymbol{\omega}^i_t,\mathbf{a}^i_{t-1},\{\boldsymbol{\delta}^j_t\}_{j\neq i}\bigr].
\end{equation}
During training, we optionally inject scaled Gaussian noise into $\mathbf{o}_t$.

\subsection{Actions}
Each agent outputs $\mathbf{a}^i_t\in[-1,1]^4$. We map to desired thrusts by
\begin{equation}
\mathbf{u}^i_t=\tfrac{\mathbf{a}^i_t+\mathbf{1}}{2}\in[0,1]^4,
\qquad
\mathbf{f}^{i,\mathrm{cmd}}_t=\mathbf{u}^i_t\,f^i_{\max}.
\end{equation}
A first-order lag on a rotor speed proxy approximates non-ideal actuation \cite{molchanov_sim--multi-real_2019}. Since thrust grows approximately with the square of rotor speed, we define
\begin{equation}
\boldsymbol{\nu}^{\,i}_t=\sqrt{\mathbf{f}^{i,\mathrm{cmd}}_t},
\quad
\tilde{\boldsymbol{\nu}}^{\,i}_{t+1}
=\tilde{\boldsymbol{\nu}}^{\,i}_{t}
+\alpha^i\!\left(\boldsymbol{\nu}^{\,i}_t-\tilde{\boldsymbol{\nu}}^{\,i}_{t}\right),
\quad
\alpha^i=\tfrac{\Delta t}{\tau^i},
\label{eq:motor-first-order}
\end{equation}
and apply thrust as
\begin{equation}
\mathbf{f}^i_t=\mathrm{clip}\!\bigl((\tilde{\boldsymbol{\nu}}^{\,i}_{t+1})^2,\,0,\,f^i_{\max}\bigr).
\end{equation}
Working in the rotor speed domain aligns the lag more closely with motor and propeller time constants. The filtered speed proxy is initialized near hover and set to zero for grounded starts.
In Mujoco the thrust of motor $j$ on vehicle $i$ is applied as an upward force along the body $z$ axis at the motor position and a reaction torque proportional to thrust is added with rotor spin sign and fixed thrust to torque coefficient $k_{\tau}=0.006$. All control runs at $250\,\mathrm{Hz}$. On the real hardware, the policy output controls \gls{pwm} directly without battery voltage compensation. We scale $\mathbf{u}^i_t$ to \gls{pwm} duty cycle. This is justified because in the operating range of the micro brushed motors, thrust is approximately proportional to duty cycle and in the model, the commanded thrust is proportional to $\mathbf{u}^i_t$. The same normalized action $\mathbf{u}^i_t$ can therefore be interpreted as a normalized thrust command in simulation and as a normalized \gls{pwm} command on hardware. Direct \gls{pwm} output avoids the typical thrust mixing step that can lead to motor saturation and therefore improves robustness, especially when operating close to the actuation limits of small lightweight quadrotors with low thrust-to-weight ratio.

\subsection{Simulation and Transition Model}
We simulate rigid body dynamics with Mujoco at $250\,\mathrm{Hz}$ with step $\Delta t=0.004\,\mathrm{s}$. At time $t$ the environment samples the next state from the transition function $s_{t+1} \sim P(\cdot \mid s_t, a_t)$ induced by one Mujoco physics step with our actuation model. 
It aggregates actuator forces and torques, tendon tension when the cable is taut, contact, and gravity. 
Each quadrotor is modeled as a freejoint rigid body and connects to the payload body through a Mujoco tendon of length $L$ that exerts tension along the line from payload to robot only when taut, and zero otherwise. Domain randomization parameters and external disturbances enter $P$ at every step. 
Contacts and friction are solved by the physics engine. This model captures multi-body coupling and the hybrid slack and taut cable modes while remaining fast enough for large scale parallel training.

\subsection{Domain Randomization for Sim-to-Real Transfer}

\begin{figure}[]
  \includegraphics[width=0.95\columnwidth]{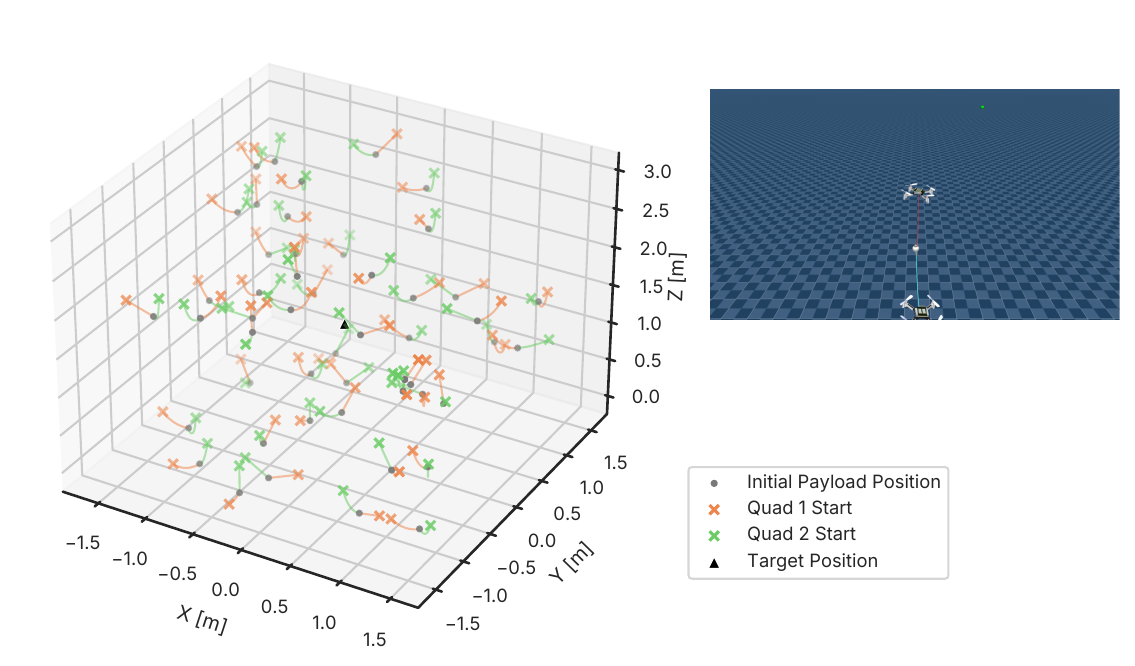}
  \caption{50 randomized initial states for $s_0\!\sim\!\rho_0$; harsh cases (slack cables, ground starts) are included. The target is at the center. the top-right shows one state in MuJoCo.}
  \label{fig:env-reset}
\end{figure}

To narrow the reality gap between simulation and reality, we rely on \gls{dr} techniques.
We introduce randomization into several aspects of our training pipeline, including randomized initial states, actuator dynamics parameters, observations, targets, and stochastic disturbances. 
Our JAX and GPU-based training setup enables training with \gls{dr} on a large number of environments, resulting in effective zero-shot transfer and robust performance in the real world.
Our \gls{dr} strategy includes the following:

\smallskip
\noindent\textbf{Initial states}: The payload is sampled around a nominal target, and quadrotors are placed on a spherical shell clipped by cable length, with randomized attitudes and linear and angular velocities. Challenging cases, such as ground starts and slack cables, are included, as shown in Fig.~\ref{fig:env-reset}.

\smallskip
\noindent\textbf{Dynamics and actuator parameters}
We randomize the per motor thrust cap around a quad level base to remain robust under battery discharge and motor aging. For each quad we draw a base thrust from $\mathcal{U}(0.105,0.15)\,\mathrm{N}$ and add an independent motor offset from $\mathcal{N}(0,0.008^2)\,\mathrm{N}$, then clip to $[0.095,0.16]\,\mathrm{N}$. The actuation lag time constants $\tau^i$ are sampled in $\mathcal{U}(0.004,0.05)\,\mathrm{s}$. The filtered thrust state receives a small perturbation at every reset and we inject occasional bounded steps in the filtered \textsc{rpm} proxy. Together, these randomizations support zero-shot sim-to-real transfer despite large variation of motor characteristics in micro brushed motors.

\smallskip
\noindent\textbf{Observations}:
We inject Gaussian noise into the
global observation vector at each timestep. Concretely, given an observation $\mathbf{o}$, we sample a
standard normal vector $\bm{\eta} \sim \mathcal{N}(0, \mathbf{I})$ of the same dimension, and compute the noisy observation using $\mathbf{o}' = \mathbf{o} + \sigma_\text{obs} \Lambda \bm{\eta}$, where $\sigma_\text{obs}$ is a tunable noise amplitude and $\Lambda$ is a diagonal scaling term to tune the noise for each observation component.

\smallskip
\noindent\textbf{Stochastic disturbances and target randomization.} At each step, we randomly apply bounded external wrenches. A random \gls{uav} receives a small force $f \sim \mathcal{U}(0,0.05)$N and a torque $\tau \sim \mathcal{U}(0,0.03)$Nm in a direction biased toward its body $z$-axis. The payload receives a small impulse force $f_p \sim \mathcal{U}(0,5)$N. We also add occasional jumps in the filtered RPM proxy, and when enabled, bounded random target updates around the goal to improve trajectory following.

\subsection{Reward Design}

\newcommand{\meanI}[1]{\avg_{i}\!\left[#1\right]}        %
\newcommand{\meanIJ}[1]{\avg_{i\neq j}\!\left[#1\right]}  %
\newcommand{\meanJ}[1]{\avg_{j}\!\left[#1\right]}         %

We propose a modular reward that applies to one or many quadrotors, with or without a cable-suspended payload. It combines tracking, stability, and safety, and uses a bounded exponential envelope to keep gradients smooth \cite{molchanov_sim--multi-real_2019,Eschmann2024,kaufmann_benchmark_2022,huang_collision_2024}. At a low level, the terms promote fast recovery with low swing and tilt, taut cables, safe spacing, and smooth thrust for sim-to-real transfer. We design reward components to lie in the range $[0,1]$ wherever possible. This limits trade-off tuning, avoids learning to terminate behavior, and removes the need for a curriculum reward.
Our composite reward is defined as  
\begin{equation}
    r = r_\mathrm{track} \, r_\mathrm{stable} + r_\mathrm{safe},
    \label{eq:reward}
\end{equation}
where $r_\mathrm{track}$ rewards small payload error and aligns payload velocity with the target direction, and $r_\mathrm{stable}$ combines terms that cap velocity smoothly, limit payload swing, keep the body-$z$ axis near vertical, and maintain taut cables. The safety component $r_\mathrm{safe}$ promotes temporal smoothness in actions, balanced motor thrust distribution, and discourages action saturation. It also encourages collision avoidance and distance keeping.
The coupling of tracking and stability in \eqref{eq:reward} encourages high speeds only when the payload is stable, while the additive safety reward applies strict incentives independent of tracking. The same reward structure scales with the number of robots $Q$, and with minor adjustments, also applies in payload-free scenarios. Full details of the sub-components of the individual rewards in \eqref{eq:reward} are provided in the Appendix.

\subsection{Training and Policy Architecture}
We train decentralized policies with \gls{ippo}~\cite{witt_is_2020} extending the JaxMARL implementation~\cite{flair2023jaxmarl}. All agents share parameters and act independently from local observations. Synchronous vectorized actors collect \(N\times T\) transitions per update and we optimize the \gls{ppo} objective with generalized advantage estimation, value loss, and an entropy bonus.

The actor is a \gls{mlp} with three hidden layers \([64,64,64]\) with \texttt{tanh} activations and a linear mean head. The action distribution is a diagonal Gaussian with learned log standard deviation. The critic is an \gls{mlp} \([128,128,128]\) that maps to a scalar value. We use orthogonal weight initialization with zero bias, sample actions during training, and use the mean at evaluation. In our experiments \gls{ippo} is sufficient for up to three quadrotors. For larger teams or stronger partial observability a centralized training variant with a joint critic such as \gls{mappo} may be beneficial. 

Hyperparameters were selected with Bayesian optimization over fixed compute budgets. While the pipeline can produce a usable policy in minutes when trained with a small number of environments, we find that with full \gls{dr} robust performance requires significant parallelization. We therefore use \(N{=}16,384\) environments, which reduces gradient variance, yields smooth learning curves, and shows little sensitivity to the random seed, while delivering very robust performance under all randomized conditions. Key settings are summarized in Table~\ref{tab:hyperparams}. Unless stated otherwise, all reported results use these defaults on a single Nvidia Ada RTX~4000 GPU with \(20\,\mathrm{GB}\) memory and \textsc{jax} and \textsc{mjx} parallelization.

\begin{table}[]
\centering
\caption{Key training and model hyperparameters.}
\label{tab:hyperparams}

\renewcommand{\arraystretch}{1.0}
\setlength{\tabcolsep}{4pt}
\resizebox{0.78\columnwidth}{!}{
\begin{tabular}{ll}
\hline
\textbf{Parameter} & \textbf{Setting} \\
\hline
Algorithm & \gls{ippo}~\cite{witt_is_2020} \\
Actor network & MLP \([64,64,64]\), \texttt{tanh} \\
Critic network & MLP \([128,128,128]\), \texttt{tanh} \\
Initialization & Orthogonal weights, zero bias \\
Action distribution & Diagonal Gaussian, learned log std \\
Rollout & \(N{=}16{,}384\) envs, \(T{=}128\) steps \\
Optimization & 256 minibatches, 8 epochs per update \\
Learning rate & \(4\times10^{-4}\) \\
Entropy coefficient & 0.01 \\
Value loss coefficient & 0.5 \\
Clip range & 0.2 \\
Grad norm clip & 0.5 \\
Discount \(\gamma\) & 0.997 \\
GAE \(\lambda\) & 0.95 \\
Episode length & 3072 steps \(\approx 12.3\,\mathrm{s}\) at 250\,Hz \\
Control rate & 250\,Hz, one sim step per action \\
Observation noise std & 1.0 \\
Action noise std & 0.0 \\
Total environment steps & \(2\times10^{9}\) \\
Seed & 0 \\
Tuning & Bayesian optimization \\
Hardware & Single Nvidia Ada RTX~4000, \(20\,\mathrm{GB}\) \\
\hline
\end{tabular}}
\end{table}

\section{Experiments and Results}
This chapter describes the evaluation methodology and outlines the metrics and figures that we used to quantify the performance of our learned policies on both single-quadrotor and multi-quadrotor cable-suspended payload transport tasks.

\subsection{Baseline Comparison}

We benchmark the decentralized \gls{rl} policy against the baseline of \cite{wahba_kinodynamic_2024}, which models each cable as a rigid rod and relies on centralized trajectory optimization with an online tracker. This limits responsiveness to cable swing, mode changes, disturbances, and payload variations. In the two quadrotors scenario, the payload begins at \((0,0,1.5)\,\mathrm{m}\) and vehicles randomized around it, both pursuing the same goal. We evaluate in simulation for statistical analysis and tightly controlled initial conditions. Each method runs \(N=1000\) trials with randomized payload states and vehicle poses. To favor the baseline we precompute a polynomial trajectory from start to target and have the baseline track it.

Our learned policy achieves \(797\) out of \(1000\) successful recoveries \(\bigl(79.7\,\%\bigr)\) with a mean speed of 0.58~m/s. The baseline achieves \(435\) successful recoveries \(\bigl(43.5\,\%\bigr)\) with a mean speed of 0.27~m/s. Given identical start and goal, the higher mean speed and the higher recovery rate indicates shorter and more reliable recoveries for our method. Qualitatively our policy damps swing quickly and follows a nearly straight approach, whereas the baseline often dips and then spirals, which is consistent with its rigid rod cable model.

\begin{figure}[]
    \centering
    \includegraphics[width=\textwidth]{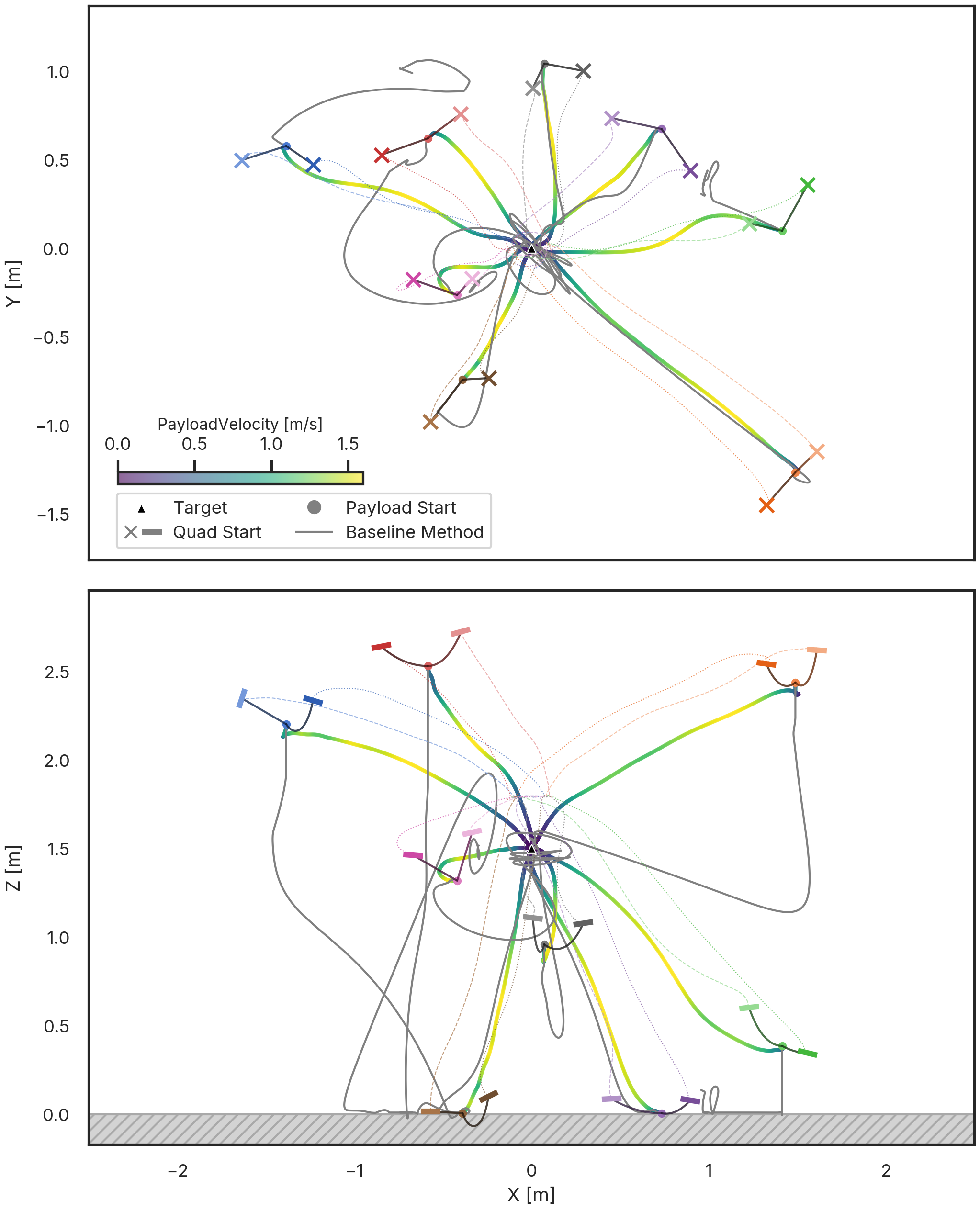}
    \caption[]{Example recovery trajectories from eight harsh initializations. Top: XY; bottom: XZ. Our Method with visualized payload velocity (colormap) vs.\ baseline (thin gray).}
    \label{fig:baseline_recovery}
\end{figure}
\subsection{Generalization}
\begin{figure}[ht]
    \centering
    
    \includegraphics[width=\textwidth]{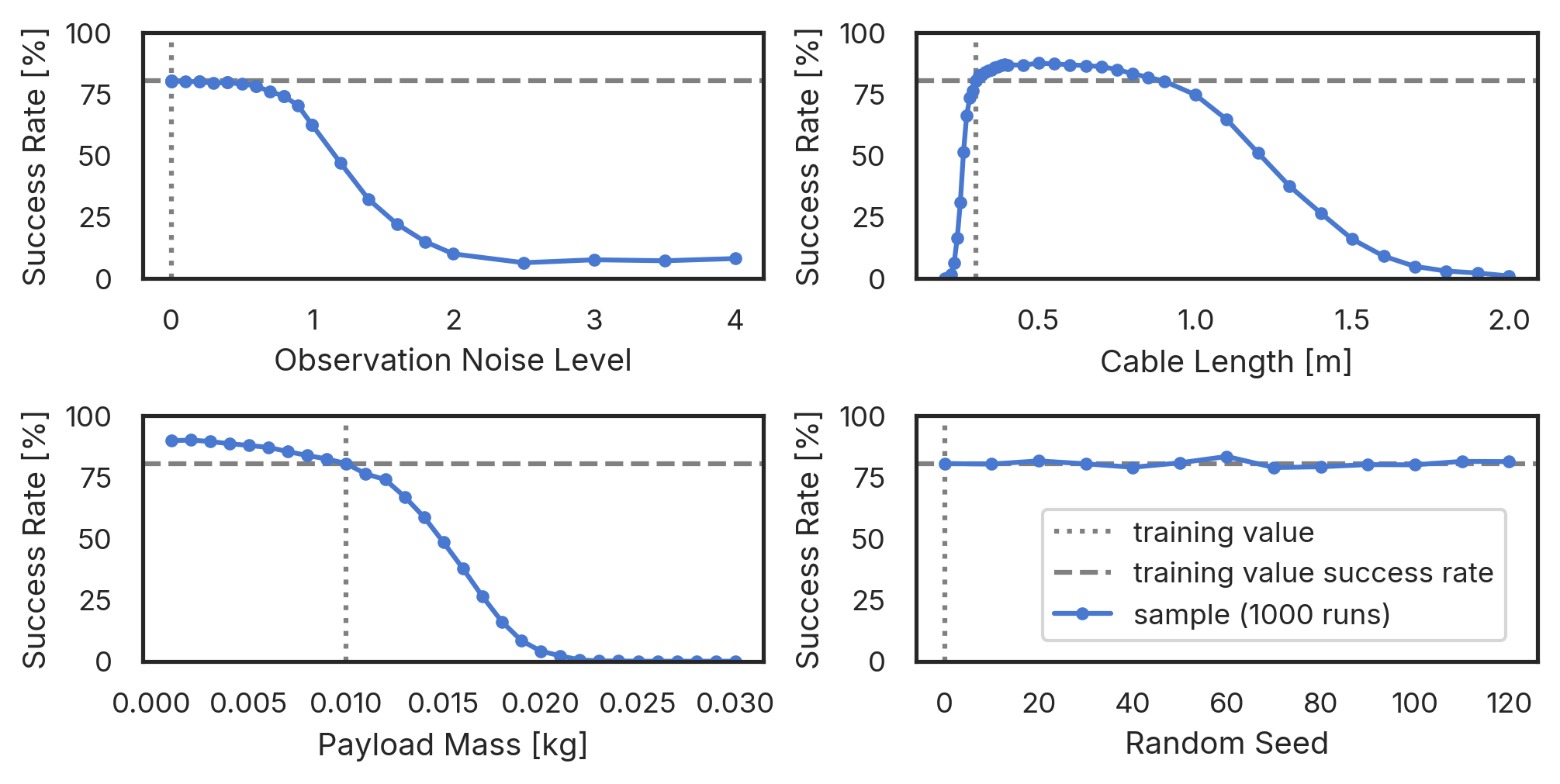}
    \caption[Policy Generalization]{Evaluation of the learned policy's generalization capabilities in the two quadrotors with payload scenario. Each datapoint represents the percentage of successfully recovered runs out of 1000 runs in an environment that only differs in the specific value adjusted. The policy is trained on a 0.3~m cable length and a payload mass of 0.01~kg.}
    \label{fig:generalization}
\end{figure}

 We assess generalization in the two quadrotor payload task by sweeping cable length ($L$), payload mass, observation noise scale ($\sigma_\text{obs}$), and initialization seed. Success is the fraction of runs that recover within 10\,s. As shown in Fig.~\ref{fig:generalization}: (i) Cable length: shorter cables sharply reduce success (increased collision risk). Moderate increases above 0.3\,m improve success, while very long cables ($>1$\,m) degrade performance as the payload becomes harder to stabilize. (ii) Payload mass: performance is robust across a broad range; lighter payloads show a slight drop, and heavier payloads reduce success more noticeably. (iii) Observation noise: the policy is tolerant up to a scale of about $1$, with a steep decline beyond that. (iv) Seeds: results are largely insensitive to initialization, consistent with training across 16k parallel environments. Overall, the policy generalizes well to unseen dynamics without explicit domain randomization, indicating strong robustness. Future work could explore randomizing payload weight, shape, and cable length to further expand the operational range.
\subsection{Scalability}
\begin{figure}[t!]
    \centering
    
    \includegraphics[width=\textwidth]{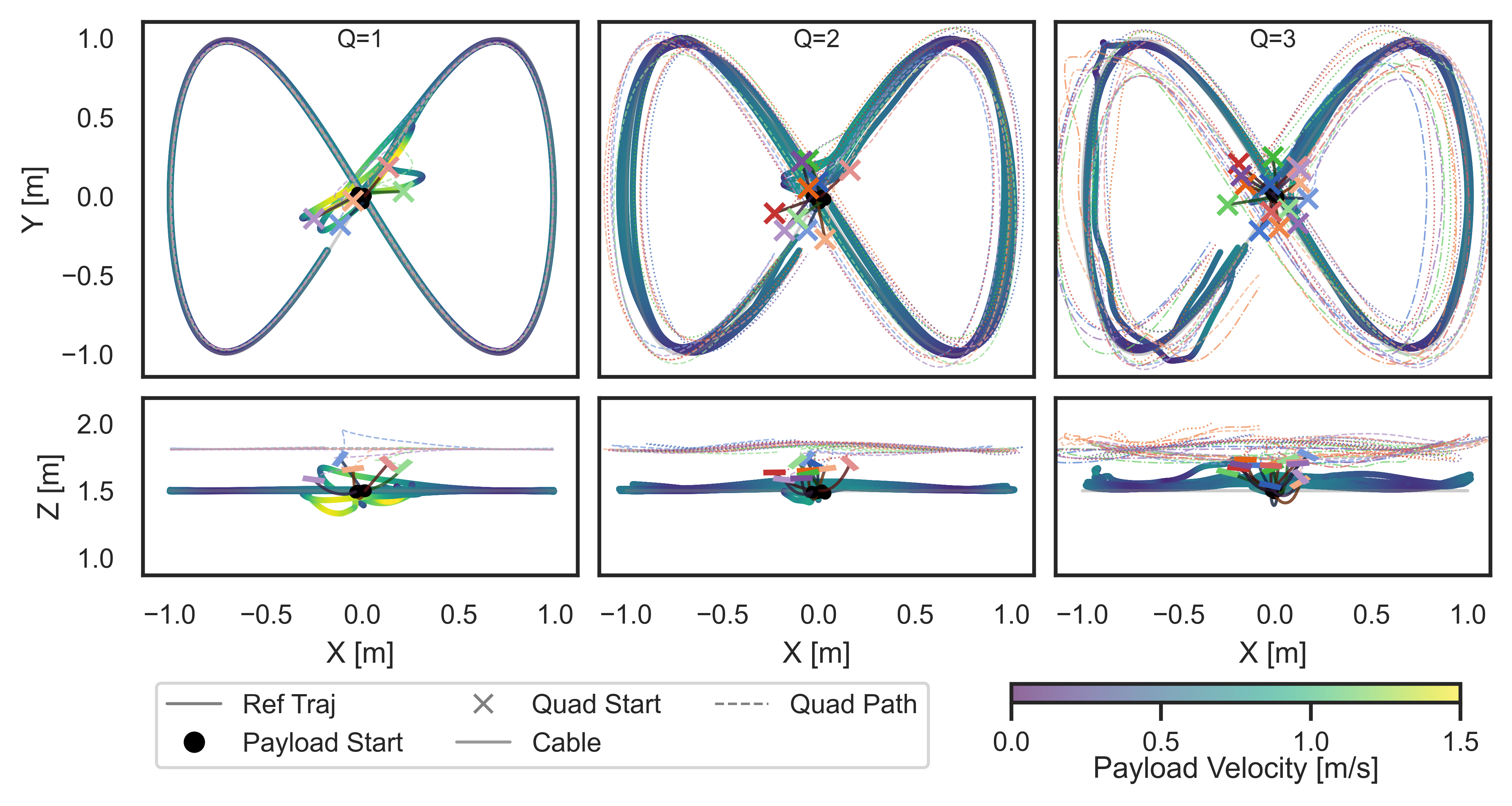}
    \caption{Figure eight trajectory tracking in simulation with one, two, and three quadrotors carrying a payload. Each column shows five runs starting at harsh initial conditions. Left $Q{=}1$ quadrotors, middle $Q{=}2$, right $Q{=}3$. The solid colormap curve is the payload trajectory with color indicating speed magnitude. Dotted curves are vehicle trajectories.}
    \label{fig:scale-8}
\end{figure}

Our framework scales from a single vehicle to larger teams. We evaluate $Q\in\{1,2,3,6\}$ on two tasks in simulation, harsh recovery from initialization as shown in Fig.~\ref{fig:env-reset} to a fixed target and tracking of a figure eight reference trajectory as shown in Fig.~\ref{fig:scale-8}. Each setting uses $1000$ trials with a policy trained for the given team size.

For $Q{=}1$ the system stabilizes from up to $1\,\mathrm{m}$ displacement in about $2\,\mathrm{s}$ with a $99\%$ success rate and tracks the figure eight with a small phase lag.  
For $Q{=}2$ recovery succeeds in $81\%$ of trials with a similar settling time and failures occur mainly at the beginning under extreme initial states that cause collisions.  
For $Q{=}3$ with a $20\,\mathrm{g}$ payload, the team reaches $60\%$ recovery success and shows coordinated load sharing.  
For $Q{=}6$ with a $40\,\mathrm{g}$ payload, most runs terminate early, which exposes a coordination limit.

The main bottleneck seems to be permutation sensitivity in peer observations. Sorting peers by distance or adding attention-based aggregation and a richer payload representation are promising next steps.
\subsection{Sim-to-Real Transfer}
We deploy the learned decentralized policy on Crazyflie quadrotors by exporting it to TFLite and compiling with STM32Cube.AI for the STM32F405. An identical policy runs fully onboard each quadrotor at \(250\,\text{Hz}\), without a centralized coordinator or cross-vehicle communication. Each robot builds its observation from the onboard \gls{ekf} state (position, velocity, attitude, and rates) together with motion-capture positions of the payload and teammates. The policy outputs individual motor commands converted to PWM and sent directly to the motors.

\begin{figure}[]
    \centering
    \includegraphics[width=\textwidth]{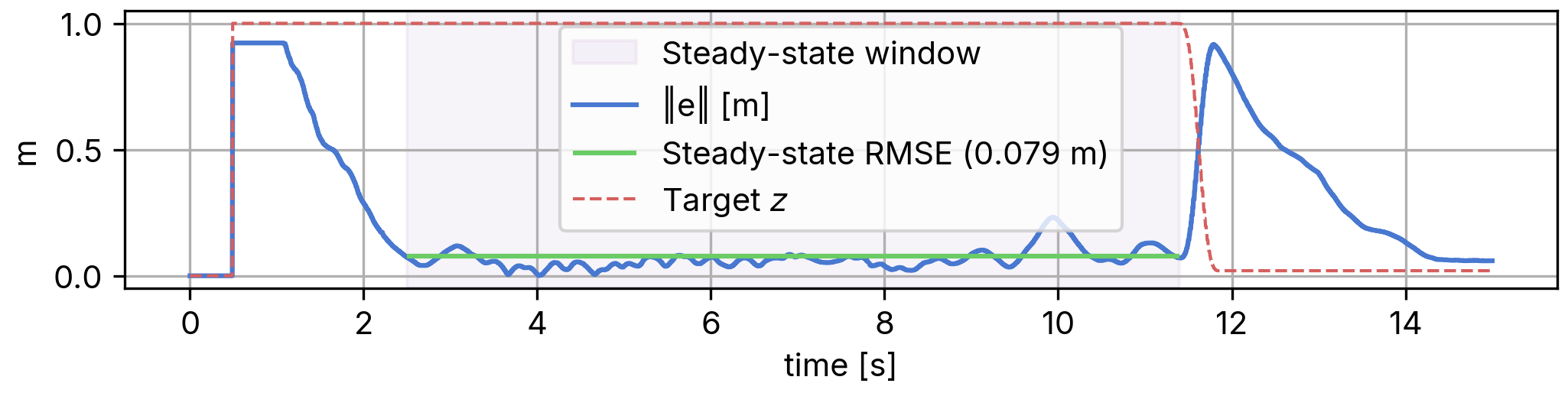}
    \caption[]{Payload position error during a 15s flight on the real hardware with takeoff and landing under wind disturbance and $10\,\mathrm{g}$ payload.}
    \label{fig:takeoff_land}
\end{figure}

\begin{figure}[]
    \centering
    
    \includegraphics[width=\textwidth]{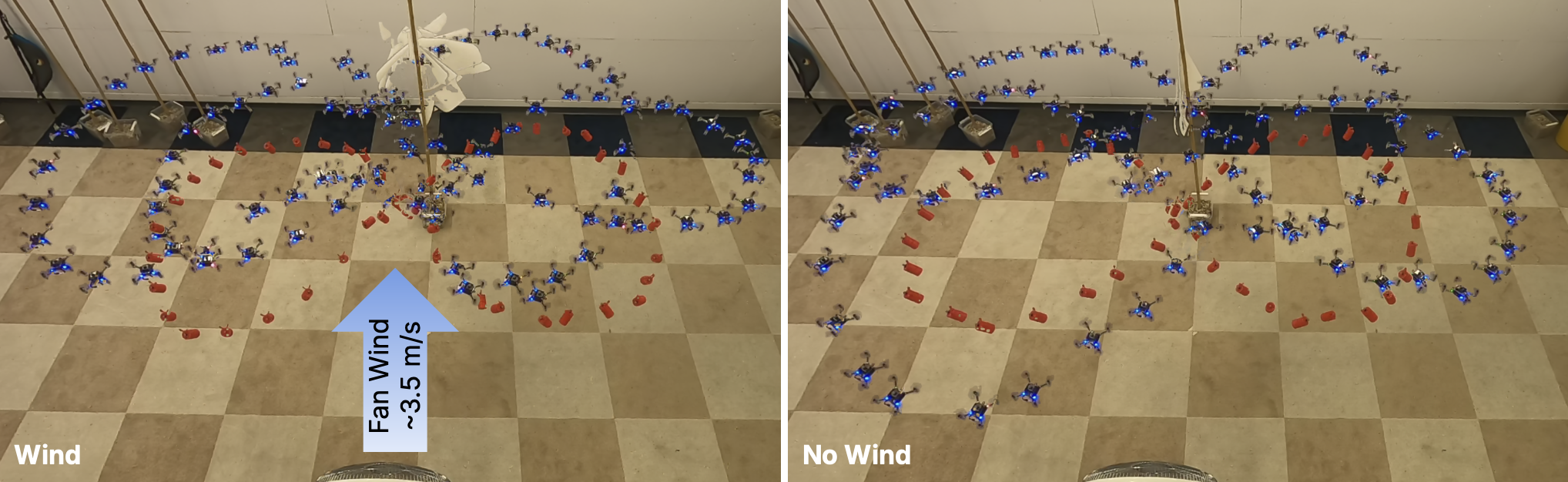}
    \caption[]{Two quadrotors with \gls{rl} policy cooperating to ensure the payload stays on a figure eight trajectory with and without wind disturbance.}
    \label{fig:fig8}
\end{figure}

Flight tests demonstrate robust autonomous takeoff, strong disturbance rejection, and stable flight in wind for a single quadrotor (with and without payload) and two quadrotors carrying a payload. In wind trials, the measured average wind speed at the target and figure-eight trajectory midpoint is $3.5\,\mathrm{m/s}$. Figures~\ref{fig:overview} and \ref{fig:takeoff_land} show agile behavior, performing a rapid takeoff from the ground to a $1\,\mathrm{m}$ altitude in $2.5\,\mathrm{s}$. Despite wind, the policy maintains a steady-state position-holding Root Mean Square Error (RMSE) of $0.077\,\mathrm{m}$, only a marginal increase over the $0.064\,\mathrm{m}$ RMSE without wind. Figure~\ref{fig:overview} also illustrates disturbance rejection with two quads and a $10\,\mathrm{g}$ payload, while Fig.~\ref{fig:fig8} demonstrates maintaining a figure-eight payload trajectory under wind.


\section{Conclusion}

We present a unified framework for training and deploying reinforcement learning controllers for single and multi-agent cable payload transport while taking into account hybrid cable dynamics. 
 Our end-to-end decentralized policy directly outputs motor-level PWM  commands at 250 Hz, enabling deployment on the resource-constrained Crazyflie 2.1 platform, which operates in our test scenarios near its actuation limits due to its low thrust-to-weight ratio. 
In both simulation and hardware, the approach achieves agile, robust transport, including zero-shot transfer with disturbance rejection and autonomous takeoff, capabilities that are unattainable with the baseline controller. 
These results highlight the potential of decentralized \gls{marl} with motor-level control as a scalable solution for cooperative aerial manipulation.

An interesting avenue for future research is to extend our work to larger multi-\gls{uav} teams by adopting order-invariant peer encodings through sorting, a centralized critic, or attention-based aggregation, enabling policies that generalize across team sizes. Additionally, integrating observation history to address partial observability and improve action smoothness, along with safety and obstacle-avoidance mechanisms, would support scalable deployment in cluttered environments.

\section*{APPENDIX}

\subsection{Reward Function Specifics}
Here we describe the components of our reward defined in equation \eqref{eq:reward} as $r = r_\mathrm{track} r_\mathrm{stable} + r_\mathrm{safe}$.
In the following, the mean over agents or motors is denoted by $\avg[\cdot]$. 
We use a reward shaping function $\Phi_{s}(x)=\exp(-s\,|x|)$ to shape reward components so that each term is bounded and the derivatives are well-behaved.
Further, let $d=\|\mathbf{e}^P\|$ and $g(d)=\min(3d,1) + c_f$, which helps to shrink the allowed payload speed near the target and keeps a small floor through the use of the scalar constant $c_f$. 

\smallskip
\noindent\textbf{Tracking}: The reward for tracking is defined as
\begin{equation}
    r_{\mathrm{track}}
  = \tfrac{1}{2}\bigl(r_{\mathrm{pos}}+r_{\rm dir}\bigr),
\end{equation}
where
\begin{align*}
  r_{\mathrm{pos}}
  &= \Phi\!\bigl(\|\mathbf{e}^P\|\bigr), \quad
  \mathbf{v}_{\rm dir}= \frac{\mathbf{v}^P}{\|\mathbf{v}^P\|+\varepsilon},\quad
  \mathbf{e}_{\rm dir}
  = \frac{\mathbf{e}^P}{\|\mathbf{e}^P\|+\varepsilon},\\
  r_{\rm dir}
  &= \Phi_{s_{\rm align}}\!\bigl(1-\mathbf{v}_{\rm dir}\!\cdot\!\mathbf{e}_{\rm dir}\bigr),\quad
  s_{\rm align}= \min\!\bigl(c_g\,\|\mathbf{e}^P\|,\,c_s\bigr).
\end{align*}
Here $r_{\mathrm{pos}}$ rewards small payload errors. The direction term aligns the payload velocity with the target direction for a linear trajectory. The factor $c_g$ yields strong guidance when far away, and the cap $c_s$ reduces sharpness near the goal.

\smallskip
\noindent\textbf{Stability}: The stability reward is defined as:
\begin{equation}
    r_{\mathrm{stable}}
  = \tfrac{1}{5}\Bigl(
      r_{\mathrm{velP}} + r_{\mathrm{velQ}} +  \lambda_{\mathrm{yaw}} r_{\mathrm{yaw}}
      + \lambda_{\mathrm{up}} r_{\mathrm{up}} + r_{\mathrm{taut}}
    \Bigr),
\end{equation}
where
\begin{align*}
  r_{\mathrm{velP}}
  &= \exp\!\left[
      -\left(\frac{\|\mathbf{v}^P\|}{c_\text{swing}\,v_{\max}\,g(d)}\right)^{c_\text{exp}}
     \right],\\
  r_{\mathrm{velQ}}
  &= \meanI{
      \exp\!\left[
        -\left(\frac{\|\mathbf{v}^i\|}{v_{\max}\,g(d)}\right)^{c_\text{exp}}
      \right]
     },\\
  r_{\mathrm{yaw}}
  &= \meanI{ \Phi(\omega^i_{z}) },\qquad
  r_{\mathrm{up}}
  = \meanI{ \Phi(\theta^i) },\\
  r_{\mathrm{taut}}
  &= \frac{1}{L}\!\left(
      \meanI{\|\mathbf{p}^i-\mathbf{p}^P\|}
      + \meanI{p^i_{z}-p^P_{z}}
     \right).
\end{align*}
The velocity term caps the speed smoothly. The exponent $c_\text{exp}$ gives a soft wall that becomes strict at the boundary. The factor $c_\text{swing}$ for the payload allows a slightly lower speed than the vehicles to limit swing. The yaw term discourages excessive yaw rate and tilt keeps the body $z$ axis near vertical. The taut term increases radial and vertical separation relative to the payload to keep cables engaged, normalized by cable length $L$.
$\omega_z^i$ denotes the yaw rate and $\theta^i$ denotes the tilt angle between each quadrotor’s body-frame z-axis and the world z-axis.
$\lambda_\text{yaw}$ and $\lambda_\text{up}$ are scaling terms.

\smallskip
\noindent\textbf{Safety}: The reward for safety is defined as 
\begin{equation}
    r_{\mathrm{safe}}
  = \tfrac{1}{5}\Bigl(
     - r_{\mathrm{coll}} - r_{\mathrm{oob}} - \lambda_{\mathrm{s}} r_{\mathrm{smooth}}
      - r_{\mathrm{energy}} + r_{\mathrm{dist}}
    \Bigr),
\end{equation}
where
\begin{align*}
  &r_{\mathrm{dist}} =
  \begin{cases}
    1, & Q = 1, \\
    \meanIJ{
      \operatorname{clip}\!\left(
        \frac{\|\mathbf{p}^i-\mathbf{p}^j\|-d_{\min}}
             {d_{\mathrm{safe}}-d_{\min}},
        0,\,1
      \right)}, & Q > 1,
  \end{cases} \\
  &r_{\mathrm{coll}}=c_\text{coll}\,\mathbb{I}_{\mathrm{coll}},\quad
  r_{\mathrm{oob}}=c_\text{oob}\,\mathbb{I}_{\mathrm{oob}},\quad
    \bar a^i_t = \meanJ{ a^i_{t,j} },\\
  &r_{\rm smooth}
  = \tfrac{1}{2}\!\left(
       \meanI{ \|\mathbf{a}^i_t-\mathbf{a}^i_{t-1}\|_{1} }
      + \meanI{ \|\mathbf{a}^i_t-\bar a^i_t\mathbf{1}\|_{1} }
     \right),\\
  &r_{\mathrm{energy}}
  = \meanI{
      \meanJ{
        \exp\!\bigl(-c_\text{b}|a^i_{t,j}|\bigr)
        + \exp\!\bigl(c_\text{b}(a^i_{t,j}-1)\bigr)
      }
     }.
\end{align*}
where $d_\text{min},d_\text{safe}$ reflect Crazyflie arm span and cable clearance,
$r_{\mathrm{coll}}$, $r_{\mathrm{oob}}$ denote binary collision and out-of-bounds indicators,
and $c_\text{coll}, c_\text{oob}$ dominate the other bounded positive terms, and strongly discourage unsafe behavior.

Here the left term in $r_{\rm smooth}$ penalizes changes between consecutive actions and enforces temporal smoothness, while the right term penalizes deviations of the four motor commands from their mean and encourages a balanced spatial thrust distribution. The energy barrier $r_{\mathrm{energy}}$ uses the coefficient $c_b$ to softly repel actions near $0$ and $1$, which discourages saturation while remaining bounded and smooth.
We use the following values of the empirically determined scalar constants:
$
s=2,
c_f=0.02, 
c_g=40,
c_s=2,
c_\text{exp} = 8,
c_\text{swing} = 0.75,
c_\text{coll} = c_\text{oob} = 10, 
c_b = 50.
$
We set $d_\text{min}$ to 0.15m, and  
$d_\text{safe}$ to 0.18m.
The coefficients $\lambda_i$ and the cap $v_{\max}$ tune the trade off between agility and caution. The rest of the reward coefficients do not require tuning. For two quadrotors we use $\lambda_{\mathrm{yaw}}{=}10$, $\lambda_{\mathrm{up}}{=}5$, $\lambda_{\mathrm{s}}{=}10$ and $v_{\max}=1.5m/s$ to get desired robust but agile behavior.

\balance

\bibliographystyle{IEEEtran}
\bibliography{literatur/bibliography}

\end{document}